\documentclass[letterpaper, 10 pt, conference]{ieeeconf}  
\let\labelindent\relax

\IEEEoverridecommandlockouts                              

\usepackage{graphicx} 
\usepackage{amsmath} 
\usepackage{amssymb}  
\usepackage{booktabs}
\usepackage{color, colortbl}
\usepackage{multirow}
\usepackage{makecell}
\usepackage{adjustbox}
\usepackage{cite}
\usepackage{wrapfig}
\usepackage{enumitem}

\definecolor{Gray}{gray}{0.8}

\title{\LARGE \bf
Safety-Critical Traffic Simulation with Adversarial Transfer of Driving Intentions
}

\author{Zherui Huang$^{1}$, Xing Gao$^{2\dagger}$, Guanjie Zheng$^{1\dagger}$, Licheng Wen$^{2}$, Xuemeng Yang$^{2}$, and Xiao Sun$^{2}$
\thanks{$^\dagger$Corresponding author. $^{1}$Shanghai Jiao Tong University, Shanghai, China. Email: \{huangzherui, gjzheng\}@sjtu.edu.cn. $^{2}$Shanghai Artificial Intelligence Laboratory, Shanghai, China. Email: \{gaoxing, wenlicheng, yangxuemeng, sunxiao\}@pjlab.org.cn. This work was performed during Zherui's internship at Shanghai Artificial Intelligence Laboratory.}
}
\begin{document}

\maketitle
\thispagestyle{empty}
\pagestyle{empty}

\begin{abstract}

Traffic simulation, complementing real-world data with a long-tail distribution, allows for effective evaluation and enhancement of the ability of autonomous vehicles to handle accident-prone scenarios. Simulating such safety-critical scenarios is nontrivial, however, from log data that are typically regular scenarios, especially in consideration of \textit{dynamic adversarial interactions} between the future motions of autonomous vehicles and surrounding traffic participants. To address it, this paper proposes an innovative and efficient strategy, termed IntSim, that explicitly decouples the driving intentions of surrounding actors from their motion planning for realistic and efficient safety-critical simulation. We formulate the adversarial transfer of driving intention as an optimization problem, facilitating extensive exploration of diverse attack behaviors and efficient solution convergence. Simultaneously, intention-conditioned motion planning benefits from powerful deep models and large-scale real-world data, permitting the simulation of realistic motion behaviors for actors. Specially, through adapting driving intentions based on environments, IntSim facilitates the flexible realization of dynamic adversarial interactions with autonomous vehicles. Finally, extensive open-loop and closed-loop experiments on real-world datasets, including nuScenes and Waymo, demonstrate that the proposed IntSim achieves state-of-the-art performance in simulating realistic safety-critical scenarios and further improves planners in handling such scenarios.

\end{abstract}

\section{Introduction}

\begin{figure*}
    \centering
    \vspace{0.08in}
    \includegraphics[width=\linewidth]{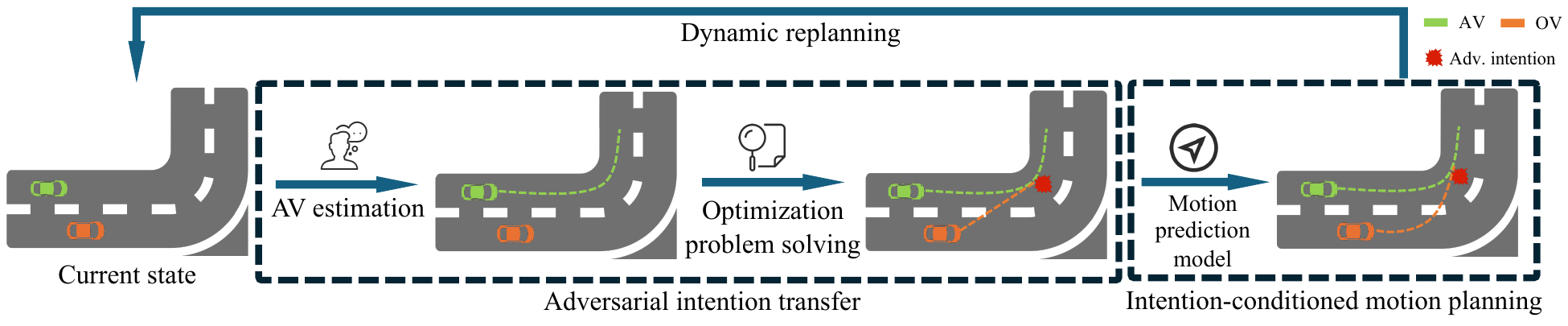}
    \caption{Illustration of the framework of IntSim. IntSim employs a decoupling strategy to realize the adversariality and realism of the generated scenarios. First, adversarial intention transfer is formulated as a constrained optimization problem in order to find the best solution to attack the autonomous vehicle directly. Subsequently, guided by the adversarial intention, IntSim leverages a motion prediction model to generate realistic trajectories.}
    \label{fig:overview}
    \vspace{-0.15in}
\end{figure*}

The evaluation and enhancement of data-driven planners \cite{planinter, end2endinter, uniad} for autonomous driving in managing corner cases, such as collision-prone scenarios, are crucial. However, these safety-critical situations are rare in the real world due to their long-tail distribution. Furthermore, they are costly to collect in consideration of the substantial costs associated with traffic accidents, which may even pose life-threatening risks.

An alternative method is to synthesize realistic safety-critical scenarios.Due to the scarcity of data, the booming deep generative technology faces challenges in direct application. Recently, several studies have explored various strategies to mitigate this issue, including adversarial optimization-based methods~\cite{advsim, strive}, reinforcement learning (RL)-based methods~\cite{feng2021intelligent, kuutti2020training}, and knowledge-based methods~\cite{ding2023causalaf, chang2023editing}. 

Nevertheless, there is room for improvement even in the state-of-the-art methods mentioned above. For example, RL-based methods, as presented in~\cite{kuutti2020training}, present strong adversarial characteristics but \textit{poor realism} and struggle to generalize to complex scenarios; adversarial optimization-based methods like STRIVE~\cite{strive}, improve on realism with the help of generative models but rely on iterative optimization in the latent space, limited by \textit{high time complexity}; Knowledge-based methods depend on prior domain knowledge, like the causal graph in \cite{ding2023causalaf}, confronted with \textit{scalability 
challenge}~\cite{cat}. 

To address these limitations, we propose IntSim (\textit{\textbf{Int}ention-aware \textbf{Sim}ulation}), an efficient and scalable approach for realistic safety-critical traffic simulation. IntSim challenges autonomous vehicles by adversarially transferring the driving intentions of surrounding actors. By explicitly decoupling the driving intentions of actors from their motion planning, we resort to two separate modules to achieve flexible and efficient attacks, along with motion planning that adheres to the constraints of realistic behavior. 

In more detail, as presented in Figure~\ref{fig:overview}, IntSim consists of two components: adversarial intention transfer and intention-conditioned motion planning. First, adversarial intention transfer is formulated as a constrained optimization problem to find the best solution to attack the autonomous vehicle directly. Specifically, the intention is characterized by the position of actors at the last future time, referred to as the goal. Subsequently, guided by the transferred intention, we adopt a goal-based motion prediction model to plan future motions of actors. Benefits from large-scale real-world data,  the motion prediction model learns from regular scenarios to generate realistic motions of actors to reach an arbitrary given goal. Thus, IntSim is able to efficiently generate rare and challenging safety-critical scenarios,  while ensuring realism.

Finally, we conduct a collection of open-loop and closed-loop experiments on real-world datasets with rule-based planners and RL-based planners to evaluate the proposed IntSim. These results demonstrate significant improvements in terms of adversariality, realism, and efficiency in safety-critical scenario simulation. Moreover, the planner trained under the safety-scritial scenarios generated by IntSim exhibits enhanced capability to handle challenging situations, successfully completing the route while avoiding collisions, and maintains its performance on regular scenarios. The main contributions of this paper are summarized below:
\begin{enumerate}[leftmargin=*]

\item We propose a novel strategy that decouples the adversarial driving intentions of surrounding actors from their motion planning for efficient and realistic simulation of safety-critical traffic scenarios. This decoupling design permits the utilization of large-scale real-world data to capture motion realism that is difficult to explicitly model.

\item We formulate adversarial intention transfer as a constrained optimization problem that is planner-agnostic and can be efficiently solved. These characteristics ensure compatibility with different motion planning algorithms and facilitate scalability for large datasets.

\item Comprehensive open-loop and closed-loop experiments demonstrate the state-of-the-art performance of IntSim in simulating safety-critical scenarios and enhancing the safety of planners accordingly.
\end{enumerate}

\section{Related Work}

\textbf{Trajectory generation and motion prediction.}
Some pioneering research~\cite{simnet,igl2022symphony,scenegen} employs end-to-end deep neural networks, like GNNs~\cite{trafficsim} and Transformer~\cite{trafficgen}, to learn to capture complex interactions in traffic scenes and generate realistic agent trajectories from large-scale real-world data. Recently, generative models like diffusion models~\cite{rempe2023trace,ctg, ctg++} are exploited to enhance the diversity of generated trajectories. However, they typically focus on regular traffic scenarios. Furthermore, a series of methods~\cite{gameformer,liu2023learning,klgame,dipp} employ game theory to model interactions between the autonomous vehicle and surrounding traffic participants to forecast their future motions for safe planning. In contrast, this work focuses on accident-prone scenario generation by controlling surrounding actors to attack the self-driving vehicle. 

\textbf{Safety-critical scenario seneration.} Existing safety-critical scenario generation methods generally fall into three categories. \textbf{(1)} Adversarial Optimization-Based Methods: AdvSim~\cite{advsim} combines random perturbations and black box searches with gradient descent optimization, while STRIVE~\cite{strive} employs conditional variational autoencoders for adversarial learning in latent space. \textbf{(2)} Reinforcement Learning-Based Methods: Koren~\cite{adaptiveStressTesting} innovatively exploits deep reinforcement learning to find near-collision situations for testing decision-making systems of autonomous driving. Feng~\cite{feng2021intelligent} proposes to use deep Q-networks to generate discrete safety-critical scenarios, whereas Kuutti~\cite{kuutti2020training} uses A2C reinforcement learning to control the surrounding vehicles of the attacked vehicle. RL-based methods do not rely on large amounts of log data but may suffer from limitations in realism and diversity. \textbf{(3)} Knowledge-Based Methods: they guide trajectory generation using prior knowledge as conditions, like safety-critical constraints and causal relationships. For instance, Diffscene~\cite{xu2023diffscene} employs safety constraints to control diffusion models, Chang~\cite{chang2023editing} and Yin~\cite{yin2021diverse} introduce a variable to model driving behavior aggressiveness, and Causalaf~\cite{ding2023causalaf} incorporates causal graphs.

Another related work CAT~\cite{cat} further takes into account the closed-loop training of planners. It exploits the multi-modal trajectory prediction model to sample multiple possible trajectories and select the one with the highest probability of colliding with the autonomous vehicle. However, the prediction model is learned from regular scenarios and fails to capture aggressive driving intention that results in collisions in corner cases. Contrary to these methods, IntSim employs a constrained optimation problem to explore diverse adversarial driving intentions and balance adversariality, realism, and efficiency with the proposed decoupling framework.

\section{Methodology}
\subsection{Problem Definition}
Given a traffic scenario, we aim to generate challenging scenarios for the autonomous vehicle (AV) by controlling the behavior of traffic actors. In line with previous works, this paper focuses on accident-prone scenarios where the AV may collide with other vehicles. We formulate a traffic scene as $(\{S^i\}_{i=1}^N, \mathcal{M})$, where $\{S^i\}_{i=1}^N$ are the trajectories of $N$ vehicles and $\mathcal{M}$ is map information. The $i$-th vehicle trajectory is represented as $S^i = \{s_0^i, s_1^i, \cdots, s_{T}^i\}$, with $s_t^i = (x_t^i, y_t^i, \theta_t^i)$ indicating 2D position and heading at timestamp $t$. The goal is to generate other vehicles' trajectories in the traffic scene $\{\hat{S}^i\}_{i \neq av}$, except for the AV $S^{av}$, in a way that leads to a collision with the AV realistically. The quality of the generated scenarios is typically measured via an objective function $\mathcal{L}$, encompassing weighted metrics such as adversariality, realism, and efficiency. Model is optimized for the  best overall performance
\begin{equation}
    \min_{\{\hat{S}^i\}_{i \neq av}} \mathcal{L} \left(\{\hat{S}^i\}_{i \neq av}; S^{av}, \mathcal{M} \right).
\end{equation}

In real world, we observe that most accidents are caused by aggressive driving behaviors. Accordingly, in this paper, we simplify situations by choosing one vehicle that exhibits aggressive driving behavior, called the opponent vehicle (OV), and interacts closely with the AV to cause it to collide. All other vehicles, referred to as background vehicles (BVs), are simulated to react to surrounding vehicles including AV, OV, and other BVs. Correspondingly, we intuitively introduce Stochastic Game~\cite{shapley1953stochastic} as our specific setting and generalize the two-player games to two-group games, where each group firstly determines their actions and then display simultaneously at each timestep, and observe the historical states and actions of all agents before choosing next actions. In games, as the team is considered a cohesive unit, before any public actions by the team, it is allowed for players within a team to strategize in order to gain a collective advantage. We set the AV as a group, and the OV and all BVs as another group. Due to the nature of traffic simulation, the OV and all BVs are controlled by us. Thus, straightforwardly, priority for action can be given to the vehicle that is most likely to collide with the AV within our group, while other vehicles tend to cooperate with its adversarial actions. The objective function is formulated as
\begin{equation}
\small
\min_{\{ \hat{S}^{bv} \} } \min_{\hat{S}^{ov}}\mathcal{L}  \left(\hat{S}^{ov}; S^{av}, \mathcal{M} \right) +  \Omega \left(\{\hat{S}^{bv}\}; S^{av}, \hat{S}^{ov}, \mathcal{M}\right),
\end{equation}
where $\hat{S}^{ov}$ and $\{\hat{S}^{bv}\}$ respectively indicate the simulated trajectories of the OV and BVs, and $\Omega$ denotes weighted metrics including realism and efficiency.

\subsection{Overview}
\label{sec:overview}
IntSim decouples safety-critical scenarios' adversariality and realism and realizes them through two stages. We find an adversarial goal position in the first stage and use the adversarial goal as guidance to generate trajectories in the second stage. In the first stage, we preliminarily find a feasible adversarial driving goal for OV that may result in a collision with AV. We achieve this by formalizing safety-critical scenarios as constrained optimization problems. In the second stage, we utilize the preliminary adversarial goal as guidance and employ a goal-based trajectory prediction model to output the final trajectory.

Generally, as shown in Figure~\ref{fig:overview}, OV generates an adversarial attack trajectory against the AV vehicle, while BV vehicles react based on the surrounding environment information. This requires IntSim to perform the following steps \emph{in turn}: (1) estimating the future trajectory of AV, (2) finding an adversarial goal position based on the estimated AV trajectory to guide OV trajectory generation, (3) generating the adversarial attack trajectory for OV, and (4) BV responding accordingly based on context information. In the following section, we will first discuss how we utilize trajectory estimation and prediction models to estimate and generate the trajectories of vehicles in the scenario, corresponding to (1), (3), (4). We then explain how to effectively determine the adversarial goal position,  leading AV to collide, as the guidance for motion generation.

\subsection{Trajectory Model}
\textbf{AV trajectory estimation.} To determine how to collide with AV, we first need to estimate its future trajectories. One straightforward solution is using a neural network (NN) to predict the trajectories of AVs. However, we find it challenging to apply NN to fit the trajectory of arbitrary AVs directly in practice. The black-box nature of planners and the scarcity of AV trajectory samples result in inaccurate predictions. Inspired by CAT~\cite{cat}, we estimate the position of the AV by sampling its trajectories in regular scenarios. Contrary to CAT, we take an arbitrary planner as a black box,  without being limited to  RL-based planners.

\textbf{OV trajectory generation.} We employ a neural network-based trajectory prediction model as the foundation to generate realistic trajectories. The model architecture is based on MTR~\cite{mtr}. The prediction model consists of an encoder-decoder architecture, taking contextual information as input and outputting future trajectories. It is trained on a large dataset of real-world data to capture complex interaction patterns and produce highly realistic trajectories. We take all vehicle trajectories $\{S^i\}$ and map information $\mathcal{M}$ as a set of polylines, where each polyline contains a set of directed points and each point is formulated by a descriptor $(x, y, \theta)$, including the 2D position coordinates and heading. Denoting the input past trajectory of the $k$-th vehicle as $S^k_{\text {in }}$, the past trajectories will be encoded as embeddings $S^k_p$ with a multilayer perceptron (MLP) 
\begin{equation}
    S^k_p = \operatorname{MLP}_\mathrm{p}\left(S^k_{\text {in }}\right).
\end{equation}
We simultaneously consider the local and global structure of the map polylines as MTR~\cite{mtr} and encode the input map polylines $M_{\text {in }}$ via two MLPs
\begin{gather}
M_{local}=\operatorname{MLP}_\mathrm{m1}\left(M_{\text {in }}\right), \quad
M_{global} = \operatorname{Pool} \left( M_{local} \right), \\
M_{p} = \operatorname{Pool}(\operatorname{MLP}_\mathrm{m2} ( [M_{local}, M_{global}] )),
\end{gather}
where $\operatorname{Pool}$ represents the max-pool operation, and $[\cdot,\cdot]$ represents concatenation operation. $M_{local}$ and $M_{global}$ respectively represent local and global structure embeddings, and $M_p$ represents encoded map embedding. Then, we leverage the multi-head attention mechanism to fuse information:
\begin{equation}
\small
    E^j= \operatorname{MultiheadAttn} \left( q=E^{j-1}, k = v=[S_p, M_p] + \operatorname{CE} \right),
\end{equation}
where $E^j$ is the output of the $j$-th attention block as well as the input of the $(j+1)$-th block, and $E^0$ is initialized as the agent encoding $S_p = [S^1_p, S^2_p, \dots, S^N_p]^T$. We introduce a learnable category embedding $\operatorname{CE}$ to encode and distinguish AV, OV, BV, and map features. Through the attention mechanism, different embeddings fully interact with each other, and the final output of attention blocks $E_p$ implicitly encapsulates all necessary information for generating future trajectories.

As mentioned in section~\ref{sec:overview}, the model needs to be capable of generating trajectories based on a given goal position. To achieve this, we establish a two-stage decoding, where the model first predicts a goal position via an MLP projection head and then completes the trajectory based on the target position. The intermediate trajectories will be decoded by encoding the given goal position and concatenating it with the context embedding $E_p$. 

\textbf{BV trajectory joint generation.} We use a transformer-based neural network to generate BV trajectories. To jointly generate BV trajectories, it needs to first encode the corresponding context for each BV and then facilitate their information interaction. For efficient computation, we tend to reuse the aforementioned well-encoded context embedding $E_p$. However, each vehicle only encodes features under its own coordinate systems, failing to interact with others. Inspired by QCNet~\cite{qcnet}, we use learnable relative position encoding to encode the relative differences between different coordinate systems. A 2D coordinate system with origin $O$ can be described uniquely by its origin coordinates and x-axis orientation, formulated as $(x_O, y_O, \theta^x_O)$, under the global systems. Given two coordinate systems, $Oxy$ and $O^{\prime}x^{\prime}y^{\prime}$. Their relationship can be described by a three-dimensional descriptor $\Delta_{Oxy \rightarrow O^{\prime}x^{\prime}y^{\prime}} = (x_O - x_{O^{\prime}}, y_O - y_{O^{\prime}}, \theta^x_O - \theta^{x^{\prime}}_{O^{\prime}})$. We encode the descriptor into a position embedding and add it to the context embedding. Thanks to the relative position embedding, we can similarly employ the attention mechanism for information interaction among BVs and, finally, decode BV trajectories jointly using an MLP projection head.

\textbf{Training.} We adopt the $\ell_1$ loss as the training loss function and apply a teacher-forcing strategy during the training of the OV model, which enforces the goal as the ground truth when predicting intermediate trajectories.

\subsection{Safety-critical Scenario Generation}
To simulate a collision with AVs, we seek adversarial intentions for OVs and complete adversarial trajectories based on the intentions. In this paper, the intention is considered to be the position of a certain goal destination. This section will elucidate how we achieve this by constructing an optimization problem and guiding the OV model in completing the trajectories.

\textbf{Notation.} We use 2D vectors to describe the position, velocity, acceleration, and jerk of agents on the x-axis and y-axis, denoting them with $\boldsymbol{p}$, $\boldsymbol{v}$, $\boldsymbol{a}$, $\boldsymbol{j}$, respectively, and denote the heading with $\theta$. We denote a certain attribution of agent $i$ at the timestamp $t$ with superscript $\cdot^i$ and subscript $\cdot_t$, \emph{e.g.}, $\boldsymbol{p}^{ov}_{t}$ representing OV's position at timestamp $t$. Besides, we use $d_{margin}(\cdot,\cdot)$ to denote the function calculating the closest distance between vehicles and road boundaries based on the input position attribution $\boldsymbol{p}$ and the map information $\mathcal{M}$ (a negative value indicates driving out of road). We denote the acceleration distribution of real-world data and generated one with $\pi_a$ and $\phi_a$, respectively. We use $p_{\pi_a}(\cdot)$ and $p_{\phi_a}(\cdot)$ to denote the probability density functions (PDF) under corresponding distributions. The distributions and PDFs of jerk are similarly denoted with $\pi_j$ and $\phi_j$, $p_{\pi_j}(\cdot)$ and $p_{\phi_j}(\cdot)$. The interval of two consecutive timestamps is represented by $\Delta t$.

\textbf{Find adversarial intention via optimization problem.} The key to scenario simulation is basically ensuring high \emph{fidelity}, particularly for safety-critical ones where \emph{adversariality} against AV needs to be taken into account. Fidelity encompasses the physical feasibility and the rationality of driving behaviors (\emph{e.g.}, drive along the center of lanes). Adversariality, on the other hand, signifies that the OV trajectory will overlap with the estimated AV trajectory at the same timestamp, resulting in a collision. As mentioned above, the optimization problem is introduced to calculate the position of an adversarial goal. To ensure that the adversarial intention is physically feasible and realistic, however, the optimization problem optimizes the \emph{whole} trajectory instead of only the position of the adversarial goal.

Mathematically, the optimization problem is formulated as:
{
\vspace{-0.07in}
\begin{subequations}
\small
\begin{align}
\max _{\phi_{a}, \phi_{j}} \quad & \mathrm{E}_{\boldsymbol{a} \sim \phi_{a}, \boldsymbol{j} \sim \phi{j}} \left[ \operatorname{log}p_{\pi_a}(\boldsymbol{a}^{ov}_{t}) + \lambda \operatorname{log}p_{\pi_j}(\boldsymbol{j}^{ov}_{t}) \right] \label{eq:a} \\
s.t. \quad & \boldsymbol{p}^{ov}_t = \boldsymbol{p}^{ov}_{t-1} + \boldsymbol{v}^{ov}_{t-1}\Delta t + \frac{1}{2}\boldsymbol{a}^{ov}_{t-1}(\Delta t)^2, & \forall t, \label{eq:b} \\
& \boldsymbol{v}^{ov}_t = \boldsymbol{v}^{ov}_{t-1} + \boldsymbol{a}^{ov}_{t-1}\Delta t, & \forall t, \label{eq:c} \\
& \boldsymbol{a}^{ov}_t = \boldsymbol{a}^{ov}_{t-1} + \boldsymbol{j}^{ov}_{t-1}\Delta t, & \forall t, \label{eq:d} \\
& \theta^{ov}_t = \operatorname{arctan}\frac{v^{ov}_{y_{t-1}}}{v^{ov}_{x_{t-1}}}, & \forall t, \label{eq:e} \\
& |\theta^{ov}_t - \theta^{ov}_{t-1}| \leq (\Delta\theta)_{max}, & \forall t, \label{eq:f} \\
& || \boldsymbol{v}^{ov}_t || \leq v_{max}, & \forall t, \label{eq:g}\\
& || \boldsymbol{a}^{ov}_t || \leq a_{max}, & \forall t, \label{eq:h}\\
& d_{margin}(\boldsymbol{p}^{ov}_t, \mathcal{M}) \geq 0, & \forall t, \label{eq:i}\\
& || \boldsymbol{p}^{av}_{t} - \boldsymbol{p}^{ov}_{t} || \leq d_{thres}, & \exists t, \label{eq:j}
\end{align}
\end{subequations}
}where $\lambda$ is the weighted hyperparameter. $v_{max}$, $a_{max}$, and $(\Delta\theta)_{max}$ denotes the physically feasible maximum velocity, acceleration, and steering angle changing within a single time interval $\Delta t$. $d_{thres}$ represents the distance threshold at which OV will collide with AV. 

In the optimization problem, the objective (Eq.~(\ref{eq:a})) is to maximize the probability of simulated kinematic quantities (acceleration and jerk) under the distribution of human-driving ones. Eqs.~(\ref{eq:b}), (\ref{eq:c}), (\ref{eq:d}), and (\ref{eq:e}) refer to the kinematic Bicycle model and abstract vehicles' movements as the Markov process, where the next state only depends on the previous state and the taken actions. To ensure the physical feasibility, Eqs.~(\ref{eq:f}), (\ref{eq:g}), and (\ref{eq:h}) set the upper bounds for velocity, acceleration, and angular velocity within a single time interval. Eq.~(\ref{eq:i}) enforces that vehicles must stay in drivable areas for rationality. Finally, Eq.~(\ref{eq:j}) makes sure the happening of the collision with the AV. 
For efficiency, the optimization problem is further reformulated into a two-level nested structure, with the inner problem being convex and efficiently solvable via standard solvers.

By solving this optimization problem, we obtain initial adversarial trajectories with the position of the adversarial goal for OV. The explicit problem formulation avoids the need for iterative inferences using deep neural network models, thereby significantly reducing computational and time costs. Moreover, by considering collisions as a constraint (Eq.~(\ref{eq:j})), the problem will not miss the possible opportunity to threaten AV and may even discover collision scenarios beyond existing data, thus enhancing adversariality. Notably, we take constrained optimization as an example in this paper to execute adversarial intention transfer and alternative strategies such as reinforcement learning can be seamlessly integrated into our framework.

\textbf{Use adversarial intention to guide trajectory generation.} While the adversarial trajectories obtained from solving the optimization problem give a feasible way to collide with AV, it does sacrifice the wealthy knowledge of driving habits embedded within large-scale real-world data. As a result, the driving trajectory only closely resembles human drivers in kinematic terms, lacking reasonable driving intentions and driving habits (as shown in the subsequent ablation study). To this end, we only utilize the final adversarial intention as the destination of the adversarial trajectory, filling in the intermediate trajectories using the aforementioned goal-based trajectory prediction model. The prediction model captures complex interactions by leveraging enormous driving data and learning from them, making the generated adversarial trajectory more realistic.

\vspace{-0.07in}
\section{Experiment}

\textbf{Experiment settings.} We aim to answer the following questions: \textit{(1) How is the quality of the generated traffic scenarios?} To answer this, we conduct comparative experiments on two real-world datasets, nuScenes~\cite{nuscenes} and Waymo~\cite{womd}, under both open-loop and closed-loop settings. \textit{(2) Can the generated scenarios help improve planners?} To answer this, we design closed-loop testing and training experiments, iterating between safety-critical scenario generation and planner training, to assess the impact of the generated scenarios on planners. \textit{(3) Are the components of our approach effective and reasonable?} To answer this, we conduct ablation studies to examine the individual contributions of each component.

For a fair comparison, we conducted experiments with existing methods on their respective reported datasets. MetaDrive~\cite{metadrive} simulation platform is adopted for the Waymo dataset.

\textbf{Planners.} We employ log replay, a rule-based planner provided by STRIVE~\cite{strive}, and an RL-based TD3 planner provided by the CAT~\cite{cat} as autonomous driving~(AV) planners. The TD3 planner is adopted in the closed-loop training. Following previous methods~\cite{strive, cat}, the planning horizon is 6s and 8s based on 2s and 1s historical horizon, and the replanning frequency is 2Hz and 10Hz, on the nuScenes and Waymo datasets, respectively.

\textbf{Metrics.} We focus on the quality and efficiency of generating safety-critical traffic scenarios. We evaluate adversariality using the AV's \textbf{collision rate}. For realism, we use \textbf{offroad rate} of the opponent vehicle (OV) to assess adherence to road structures. Additionally, OV's \textbf{global offroad rate}, which considers entire generated trajectories instead of truncating them upon collision, is introduced to indicate the reasonableness of driving intentions. We calculate the Wasserstein distribution as the \textbf{distribution distance} between generated and real-world OV kinematics. The \textbf{inroad and collision rate} measures OVs that attack without entering non-drivable areas, balancing realism and adversariality. We also report average \textbf{generation time} \emph{per step} over scenarios for efficiency evaluation. In closed-loop planner training and testing, we assess the AV's \textbf{collision} and \textbf{offroad rates} to evaluate safety handling. The \textbf{route completion} metric measures the percentage of completed navigation paths to assess planner progress.

\begin{table}
    \centering
    \vspace{0.08in}
    \small
    \caption{Safety-critical scenario evaluation on the nuScenes dataset.}
    \label{tab:openloop-nuscenes}
    \resizebox{\linewidth}{!}{
     \setlength{\tabcolsep}{1mm}{
    \begin{tabular}
    {p{1.4cm}<{\centering}|p{1.1cm}<{\centering}|p{.7cm}<{\centering}p{.8cm}<{\centering}p{.8cm}<{\centering}p{1.8cm}<{\centering}|p{.8cm}<{\centering}p{.8cm}<{\centering}|p{.8cm}<{\centering}}
    \toprule
         \multirow{2}{*}{Planner} & \multirow{2}{*}{Method} & {\scriptsize Coll. \qquad $\uparrow$ } & {\scriptsize Off~Road    $\downarrow$}& {\scriptsize  Off~Road (Global)~$\downarrow$} &{\scriptsize In-Road \&~Coll. \quad $\uparrow$ } & {\scriptsize Accel. Dist. $\downarrow$} & {\scriptsize Jerk Dist. $\downarrow$ } & {\scriptsize Time (s)~$\downarrow$}\\
        \midrule
        \multirow{3}{*}{Playback} & {\small Raw} & 0.0 & 9.4 & 9.4 & 0.0  (-) & 0.00 & 0.00 & -\\ 
         & {\small STRIVE} & 38.2 & \textbf{9.9} &  14.2 & 34.9 (91.4\%) & 0.23 & 0.82 & 14.05 \\
         & {\small Ours} & \textbf{40.6} & \textbf{9.9} & \textbf{10.4} & \textbf{38.2 (94.1\%)} & \textbf{0.10} & \textbf{0.35} & \textbf{0.71} \\
        \midrule
        \multirow{3}{*}{\makecell{Rule-based\\Planner}} & {\small Raw} & 1.6 & 8.1 & 8.1 & 1.6 (-) & 0.00 & 0.00 & - \\ 
        &{\small STRIVE} & 24.0 & 17.8 & 20.2 & 20.9 (87.1\%) & 0.27 & \textbf{0.45} & 27.27 \\
        &  {\small Ours} & \textbf{26.4} & \textbf{10.9} & \textbf{11.2} & \textbf{26.0 (98.5\%)} & \textbf{0.22} & 0.50 & \textbf{1.17} \\
        \bottomrule
    \end{tabular}
    }}
    \centering
    \small
    \caption{Safety-critical scenario evaluation on the Waymo dataset.}
    \resizebox{\linewidth}{!}{
    \setlength{\tabcolsep}{1mm}{
    \begin{tabular}
    {p{1.2cm}<{\centering}|p{1cm}<{\centering}|p{.7cm}<{\centering}p{.8cm}<{\centering}p{.8cm}<{\centering}p{1.8cm}<{\centering}|p{.8cm}<{\centering}p{.8cm}<{\centering}|p{.8cm}<{\centering}}
    \toprule
         \multirow{2}{*}{Planner} & \multirow{2}{*}{Method} & {\scriptsize Coll. \qquad $\uparrow$ } & {\scriptsize Off~Road    $\downarrow$}& {\scriptsize  Off~Road (Global)~$\downarrow$} &{\scriptsize In-Road \&~Coll. \quad $\uparrow$ } & {\scriptsize Accel. Dist. $\downarrow$} & {\scriptsize Jerk  Dist. $\downarrow$ } & {\scriptsize Time (s)~$\downarrow$ }\\
        \midrule
        \multirow{3}{*}{Playback} & Raw & 0.0 & 11.0 & 11.0 & 0.0 (-) & 0.00 & 0.00 & - \\ 
         & CAT & 89.8 & 11.6 & 21.6 & 81.4 (90.6\%) & 0.20 & \textbf{0.30} & 1.66 \\
         & Ours & \textbf{90.0} & \textbf{8.8} & \textbf{10.2} & \textbf{85.2 (94.8\%)} & \textbf{0.12} & 0.37 & \textbf{1.59} \\
        \midrule
        \multirow{3}{*}{\makecell{RL-based \\Planner}} & Raw & 19.0 & 7.6 & 11.0 & 18.4 (-) & 0.00 & 0.00 & - \\ 
        & CAT & 36.7 & 9.6 & 20.0 & 34.7 (94.6\%) & 0.26 & 0.25 & 1.85 \\
        & Ours & \textbf{45.3} & \textbf{5.4} & \textbf{7.8} & \textbf{45.1 (99.6\%)} & \textbf{0.17} & \textbf{0.13} & \textbf{1.59} \\
        \bottomrule
    \end{tabular}
    }}\label{tab:openloop-waymo}
\vspace{-0.15in}
\end{table}

\begin{figure}
    \centering
    \vspace{0.08in}
    \includegraphics[width=\linewidth]{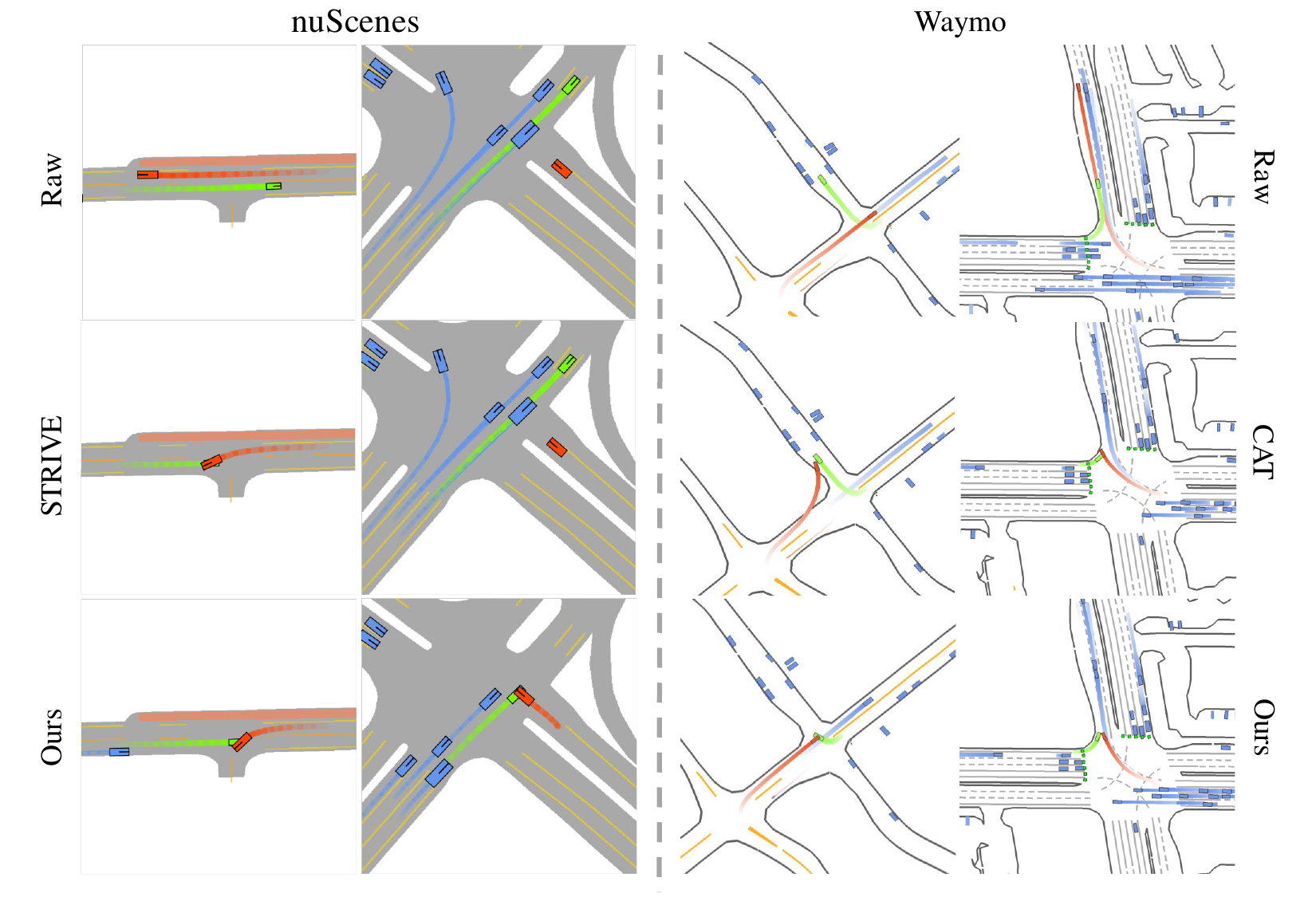}
    \caption{Visual comparison of generated scenarios on nuScenes and Waymo datasets. \textcolor{green}{AV}, \textcolor{red}{OV}, and \textcolor{blue}{BV} are colored in green, red, and blue, respectively.}
    \label{fig:qualitative}
    \vspace{-0.25in}
\end{figure}

\subsection{Evaluation for Generated Scenarios}

\textbf{Quantitative results.} Tables~\ref{tab:openloop-nuscenes} and \ref{tab:openloop-waymo} present the quantitative comparison results of safety-critical scenario generation between IntSim and the baselines. The results show that IntSim generates safety-critical traffic scenarios with higher collision rates than baseline methods while outperforming them in terms of realism, thus achieving comprehensively better performance on both datasets. The results suggest that IntSim is capable of generating challenging safety-critical scenarios for autonomous vehicles while adhering to road structures and ensuring reasonable driving intentions. Furthermore, IntSim achieves competitive performance with an average replanning time of 1.59 seconds, compared to CAT’s 1.66 seconds with $N=5$\footnote{$N$ represents the number of AV trajectories collected by CAT to estimate AV's behavior, and $N=5$ as described in its public paper.}, and certain advantages in practical applications since superior simulation quality.

\textbf{Qualitative results}. In the visual comparisons presented in Figure~\ref{fig:qualitative}, IntSim outperforms state-of-the-art baselines across two datasets. Notably, IntSim successfully generates adversarial trajectories in scenarios where the baselines fail to collide with autonomous vehicles (AVs), as illustrated in the second column of Figure~\ref{fig:qualitative}. Furthermore, our generated trajectories remain within lanes, exhibiting better realism. 
Additionally, IntSim produces intention-reasonable trajectories, in contrast to other methods that may lead to driving out of bounds unless a collision occurs (referred to as \textit{global offroad} in this paper),  as shown in the last column. The key lies in IntSim’s incorporation of explicit transfer of driving intention via optimization and intention-conditioned planning models for realistic trajectory complementation, rather than mere searching in the trajectory space or latent space of motion generation models.

\subsection{Benifit Planners via Closed-loop Training}
Table~\ref{tab:closeloop} showcases the performance comparison of the planner after closed-loop training. The mean and standard deviation of results are reported over three different seeds. We utilize the TD3 RL-based planner implemented by CAT~\cite{cat}. The planner trained on playback demonstrates better route completion but struggles to handle safety-critical near-collision scenarios. In contrast, the planner trained using safety-critical scenario generation approaches prioritizes safety and avoids potential collisions, resulting in more cautious driving behavior at the cost of lower route completion. Thanks to the balanced realism and sampling efficiency of scenarios generated by IntSim, the planner trained using IntSim achieves much better results than that with CAT in terms of collision rate and out-of-road rate while sacrificing fewer route completions.

\begin{table}[t]
    \centering
    \vspace{0.08in}
    \caption{Closed-loop evaluation of generated safety-critical scenarios on the  Waymo dataset.}
    \resizebox{\linewidth}{!}{
    \setlength{\tabcolsep}{2mm}{
    \begin{tabular}{c|p{1.1cm}<{\centering}|p{1.6cm}<{\centering}p{1.6cm}<{\centering}p{1.6cm}<{\centering}}
    \toprule
        \multirow{2}{*}{Eval.}  & \multirow{2}{*}{Train} & \multirow{2}{*}{Coll.~$\downarrow$}&Ego \qquad \qquad Off Road~$\downarrow$ & Route Completion~$\uparrow$ \\
        \midrule
        \multirow{3}{*}{Playback} & Playback & \underline{15.67 $\pm$ 3.09} & \textbf{28.00 $\pm$ 3.27} & \textbf{77.50 $\pm$ 3.25} \\
        & CAT & 17.00 $\pm$ 5.66 & 37.67 $\pm$ 1.89 & 66.03 $\pm$ 4.57 \\
        & IntSim & \cellcolor{Gray}\textbf{13.67 $\pm$ 0.94} & \cellcolor{Gray}\underline{33.67 $\pm$ 6.80} & \cellcolor{Gray}\underline{73.73 $\pm$ 5.13} \\
        \midrule
        \multirow{3}{*}{\makecell{CAT}} & Playback & 43.00 $\pm$ 7.26 & \textbf{21.33 $\pm$ 7.93} & \underline{67.77 $\pm $4.85} \\
        & CAT & \textbf{32.67 $\pm$ 6.32} & 38.67 $\pm$ 4.19 & 57.47 $\pm$ 4.93 \\
        & IntSim & \cellcolor{Gray}\underline{34.67 $\pm$ 4.78} & \cellcolor{Gray}\underline{28.33 $\pm$ 6.34} & \cellcolor{Gray}\textbf{66.53 $\pm$ 4.44} \\
        \midrule
        \multirow{3}{*}{\makecell{IntSim}} & Playback & \underline{34.00 $\pm$ 2.16} & \textbf{24.33 $\pm$ 4.03} & \underline{67.17 $\pm$ 2.95} \\
        & CAT & 37.00 $\pm$ 7.12 & 33.00 $\pm$ 0.82 & 56.27 $\pm$ 5.79 \\
        & IntSim & \cellcolor{Gray}\textbf{26.33 $\pm$ 2.87} & \cellcolor{Gray}\underline{32.00 $\pm$ 6.48} & \cellcolor{Gray}\textbf{67.63 $\pm$ 6.51} \\
        \bottomrule
    \end{tabular}}
    }
\vspace{-0.25in}
    \label{tab:closeloop}
\end{table}

\subsection{Ablation Study}
\begin{figure}
    \centering
    \includegraphics[width=\linewidth]{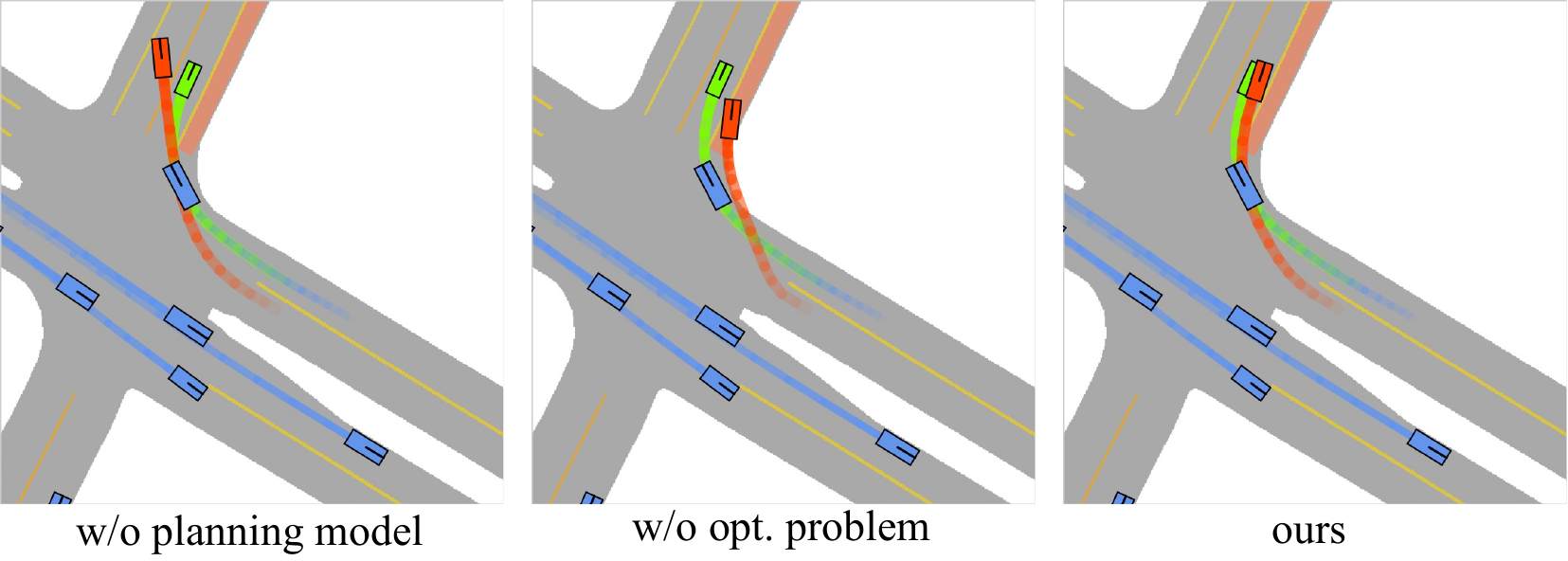}
    \vspace{-0.5cm}
    \caption{Qualitative results of ablation study on the motion planning model and the optimization problem. \textcolor{green}{AV}, \textcolor{red}{OV}, and \textcolor{blue}{BV} are colored in green, red, and blue, respectively.}
    \label{fig:abl}
    \vspace{-0.25in}
\end{figure}

\textbf{Studies on driving intention transfer.} As presented in Table~\ref{tab:abl}, the model without adversarial transfer of driving intention (ID~1) is hardly to attck AV to cause collision. With a heuristic method to obtain the adversarial goal (ID~3), interpolation between a randomly sampled point of estimated AV trajectories and OV's current position, the collision rate is largely increased, albeit at the expense of realism. It is faced with severe offroad (20.3\%), unusual (typically higher) acceleration (1.91), and jerk (6.07), and other unrealistic behaviors due to unreasonable guided goal positions as demonstrated in Figure~\ref{fig:abl}. In contrast, the proposed intention transfer via optimization effectively balances the attack success rate and the realism of generated trajectories.

\begin{table}
    \centering
    \vspace{0.08in}
    \caption{Ablation studies on the prediction model and driving intention transfer on nuScenes.}
    \label{tab:abl}
    \resizebox{\linewidth}{!}{
        \setlength{\tabcolsep}{1.5mm}{
        \begin{tabular}{c|cc|ccccc}
        \toprule
         ID & {\scriptsize \makecell{Pred.\\model}} & {\scriptsize \makecell{Intent.\\Transfer}} &{\scriptsize Coll. $\uparrow$} & {\scriptsize Offroad $\downarrow$} & {\scriptsize \makecell{Global\\offroad} $\downarrow$} & {\scriptsize \makecell{Accel.\\dist.} $\downarrow$ } & {\scriptsize \makecell{Jerk\\dist.} $\downarrow$} \\
        \midrule
         1& $\checkmark$ &   & \textbf{1.4} & \textbf{9.0} & \textbf{9.0} & 0.19 & 0.79  \\
         2& &  opt. & 45.6 & 9.9 & 16.0 & \textbf{0.05} & 0.81  \\
         3& $\checkmark$ &interp.&  75.9 & 20.3 & 24.1 & 1.91 & 6.07 \\
         \cellcolor{Gray}4& \cellcolor{Gray}$\checkmark$ & \cellcolor{Gray}opt. & \cellcolor{Gray}40.6 & \cellcolor{Gray}9.9 & \cellcolor{Gray}\textbf{10.4} & \cellcolor{Gray}0.10 & \cellcolor{Gray}\textbf{0.35} \\
         \bottomrule
        \end{tabular}
    }}
\vspace{-0.1in}
\end{table}

\textbf{Studies on intention-conditioned motion planning model.} The variant (ID 2) without the intention-conditioned motion planning model directly obtains the trajectories by solving the optimization problem.  It exhibits a higher global offroad rate (16.0\%), indicating a lack of reasonable driving habits. Moreover, it weakens the physical feasibility of the vehicle's trajectory, as illustrated in Figure~\ref{fig:abl}.

\begin{wraptable}{r}{3.2cm}
\vspace{-0.3cm}
\caption{Evaluation of motion planning model.
}
    \centering
    \resizebox{\linewidth}{!}{
    \begin{tabular}{c|cc}
        \toprule
        & ADE~(m) & FDE~(m) \\
        \midrule
        STRIVE & 0.28 & 0.45 \\
        IntSim & 0.28 & 0.41 \\
        IntSim* & 0.20 & 0.00 \\
        \bottomrule
    \end{tabular}
    }
    \label{tab:motion-model}
\end{wraptable}

The motion planning model's performance is assessed against STRIVE's generation model in trajectory forecasting over a 2-second horizon, aligning with the planning horizon of the OV, on our nuScenes test scenarios, as shown in Table~\ref{tab:motion-model}. We further evaluate the model to capture intention given future goal positions, corresponding IntSim*. The model shows competitive performance in terms of ADE and FDE, and further reduces the ADE to 0.20m when goals are provided, demonstrating effective intention-based planning.

\section{Conclusion and Future Work}
This paper proposes a novel method for generating traffic safety-critical scenarios, considering adversarial and realistic aspects through a two-stage process. We transform safety-critical scenarios into optimization problems and introduce a goal-based trajectory prediction model to generate adversarial and realistic trajectories. Both open-loop and closed-loop evaluation on real-world datasets demonstrate our advantages regarding adversariality and realism. The closed-loop training for the planner shows that our method can provide better safety assurance for autonomous vehicles. In this work, IntSim considers all agents in traffic scenes and their interaction. However, it focuses on collisions between two agents, while more complex traffic accidents involving multiple agents are limitedly considered. It is worth exploring how to generate more complex safety-critical scenarios to increase the diversity of generated data and further enhance the comprehensive development of autonomous driving planning in future work.

\bibliographystyle{IEEEtran}
\bibliography{bibfile}

\begin{thebibliography}{10}
\providecommand{\url}[1]{#1}
\csname url@rmstyle\endcsname
\providecommand{\newblock}{\relax}
\providecommand{\bibinfo}[2]{#2}
\providecommand\BIBentrySTDinterwordspacing{\spaceskip=0pt\relax}
\providecommand\BIBentryALTinterwordstretchfactor{4}
\providecommand\BIBentryALTinterwordspacing{\spaceskip=\fontdimen2\font plus
\BIBentryALTinterwordstretchfactor\fontdimen3\font minus \fontdimen4\font\relax}
\providecommand\BIBforeignlanguage[2]{{%
\expandafter\ifx\csname l@#1\endcsname\relax
\typeout{** WARNING: IEEEtran.bst: No hyphenation pattern has been}%
\typeout{** loaded for the language `#1'. Using the pattern for}%
\typeout{** the default language instead.}%
\else
\language=\csname l@#1\endcsname
\fi
#2}}

\bibitem{planinter}
A.~Sadat, S.~Casas, M.~Ren, X.~Wu, P.~Dhawan, and R.~Urtasun, ``Perceive, predict, and plan: Safe motion planning through interpretable semantic representations,'' in \emph{Computer Vision--ECCV 2020: 16th European Conference, Glasgow, UK, August 23--28, 2020, Proceedings, Part XXIII 16}.\hskip 1em plus 0.5em minus 0.4em\relax Springer, 2020, pp. 414--430.

\bibitem{end2endinter}
W.~Zeng, W.~Luo, S.~Suo, A.~Sadat, B.~Yang, S.~Casas, and R.~Urtasun, ``End-to-end interpretable neural motion planner,'' in \emph{Proceedings of the IEEE/CVF Conference on Computer Vision and Pattern Recognition (CVPR)}, 2019, pp. 8660--8669.

\bibitem{uniad}
Y.~Hu, J.~Yang, L.~Chen, K.~Li, C.~Sima, X.~Zhu, S.~Chai, S.~Du, T.~Lin, W.~Wang, \emph{et~al.}, ``Planning-oriented autonomous driving,'' in \emph{Proceedings of the IEEE/CVF Conference on Computer Vision and Pattern Recognition (CVPR)}, 2023, pp. 17\,853--17\,862.

\bibitem{advsim}
J.~Wang, A.~Pun, J.~Tu, S.~Manivasagam, A.~Sadat, S.~Casas, M.~Ren, and R.~Urtasun, ``Advsim: Generating safety-critical scenarios for self-driving vehicles,'' in \emph{Proceedings of the IEEE/CVF Conference on Computer Vision and Pattern Recognition (CVPR)}, 2021, pp. 9909--9918.

\bibitem{strive}
D.~Rempe, J.~Philion, L.~J. Guibas, S.~Fidler, and O.~Litany, ``Generating useful accident-prone driving scenarios via a learned traffic prior,'' in \emph{Proceedings of the IEEE/CVF Conference on Computer Vision and Pattern Recognition (CVPR)}, 2022, pp. 17\,305--17\,315.

\bibitem{feng2021intelligent}
S.~Feng, X.~Yan, H.~Sun, Y.~Feng, and H.~X. Liu, ``Intelligent driving intelligence test for autonomous vehicles with naturalistic and adversarial environment,'' \emph{Nature communications}, vol.~12, no.~1, p. 748, 2021.

\bibitem{kuutti2020training}
S.~Kuutti, S.~Fallah, and R.~Bowden, ``Training adversarial agents to exploit weaknesses in deep control policies,'' in \emph{2020 IEEE International Conference on Robotics and Automation (ICRA)}.\hskip 1em plus 0.5em minus 0.4em\relax IEEE, 2020, pp. 108--114.

\bibitem{ding2023causalaf}
W.~Ding, H.~Lin, B.~Li, and D.~Zhao, ``Causalaf: Causal autoregressive flow for safety-critical driving scenario generation,'' in \emph{Conference on robot learning (CoRL)}.\hskip 1em plus 0.5em minus 0.4em\relax PMLR, 2023, pp. 812--823.

\bibitem{chang2023editing}
W.-J. Chang, C.~Tang, C.~Li, Y.~Hu, M.~Tomizuka, and W.~Zhan, ``Editing driver character: Socially-controllable behavior generation for interactive traffic simulation,'' \emph{IEEE Robotics and Automation Letters}, 2023.

\bibitem{cat}
L.~Zhang, Z.~Peng, Q.~Li, and B.~Zhou, ``Cat: Closed-loop adversarial training for safe end-to-end driving,'' in \emph{Conference on Robot Learning (CoRL)}.\hskip 1em plus 0.5em minus 0.4em\relax PMLR, 2023, pp. 2357--2372.

\bibitem{simnet}
L.~Bergamini, Y.~Ye, O.~Scheel, L.~Chen, C.~Hu, L.~Del~Pero, B.~Osi{\'n}ski, H.~Grimmett, and P.~Ondruska, ``Simnet: Learning reactive self-driving simulations from real-world observations,'' in \emph{2021 IEEE International Conference on Robotics and Automation (ICRA)}.\hskip 1em plus 0.5em minus 0.4em\relax IEEE, 2021, pp. 5119--5125.

\bibitem{igl2022symphony}
M.~Igl, D.~Kim, A.~Kuefler, P.~Mougin, P.~Shah, K.~Shiarlis, D.~Anguelov, M.~Palatucci, B.~White, and S.~Whiteson, ``Symphony: Learning realistic and diverse agents for autonomous driving simulation,'' in \emph{2022 International Conference on Robotics and Automation (ICRA)}.\hskip 1em plus 0.5em minus 0.4em\relax IEEE, 2022, pp. 2445--2451.

\bibitem{scenegen}
S.~Tan, K.~Wong, S.~Wang, S.~Manivasagam, M.~Ren, and R.~Urtasun, ``Scenegen: Learning to generate realistic traffic scenes,'' in \emph{Proceedings of the IEEE/CVF Conference on Computer Vision and Pattern Recognition (CVPR)}, 2021, pp. 892--901.

\bibitem{trafficsim}
S.~Suo, S.~Regalado, S.~Casas, and R.~Urtasun, ``Trafficsim: Learning to simulate realistic multi-agent behaviors,'' in \emph{Proceedings of the IEEE/CVF Conference on Computer Vision and Pattern Recognition}, 2021, pp. 10\,400--10\,409.

\bibitem{trafficgen}
L.~Feng, Q.~Li, Z.~Peng, S.~Tan, and B.~Zhou, ``Trafficgen: Learning to generate diverse and realistic traffic scenarios,'' in \emph{2023 IEEE International Conference on Robotics and Automation (ICRA)}.\hskip 1em plus 0.5em minus 0.4em\relax IEEE, 2023, pp. 3567--3575.

\bibitem{rempe2023trace}
D.~Rempe, Z.~Luo, X.~Bin~Peng, Y.~Yuan, K.~Kitani, K.~Kreis, S.~Fidler, and O.~Litany, ``Trace and pace: Controllable pedestrian animation via guided trajectory diffusion,'' in \emph{Proceedings of the IEEE/CVF Conference on Computer Vision and Pattern Recognition (CVPR)}, 2023, pp. 13\,756--13\,766.

\bibitem{ctg}
Z.~Zhong, D.~Rempe, D.~Xu, Y.~Chen, S.~Veer, T.~Che, B.~Ray, and M.~Pavone, ``Guided conditional diffusion for controllable traffic simulation,'' in \emph{2023 IEEE International Conference on Robotics and Automation (ICRA)}.\hskip 1em plus 0.5em minus 0.4em\relax IEEE, 2023, pp. 3560--3566.

\bibitem{ctg++}
Z.~Zhong, D.~Rempe, Y.~Chen, B.~Ivanovic, Y.~Cao, D.~Xu, M.~Pavone, and B.~Ray, ``Language-guided traffic simulation via scene-level diffusion,'' in \emph{Conference on Robot Learning (CoRL)}.\hskip 1em plus 0.5em minus 0.4em\relax PMLR, 2023, pp. 144--177.

\bibitem{gameformer}
Z.~Huang, H.~Liu, and C.~Lv, ``Gameformer: Game-theoretic modeling and learning of transformer-based interactive prediction and planning for autonomous driving,'' in \emph{Proceedings of the IEEE/CVF International Conference on Computer Vision (CVPR)}, 2023, pp. 3903--3913.

\bibitem{liu2023learning}
X.~Liu, L.~Peters, and J.~Alonso-Mora, ``Learning to play trajectory games against opponents with unknown objectives,'' \emph{IEEE Robotics and Automation Letters (RA-L)}, vol.~8, no.~7, pp. 4139--4146, 2023.

\bibitem{klgame}
J.~Lidard, H.~Hu, A.~Hancock, Z.~Zhang, A.~G. Contreras, V.~Modi, J.~DeCastro, D.~Gopinath, G.~Rosman, N.~Leonard, \emph{et~al.}, ``Blending data-driven priors in dynamic games,'' \emph{arXiv preprint arXiv:2402.14174}, 2024.

\bibitem{dipp}
Z.~Huang, H.~Liu, J.~Wu, and C.~Lv, ``Differentiable integrated motion prediction and planning with learnable cost function for autonomous driving,'' \emph{IEEE transactions on neural networks and learning systems}, 2023.

\bibitem{adaptiveStressTesting}
M.~Koren, S.~Alsaif, R.~Lee, and M.~J. Kochenderfer, ``Adaptive stress testing for autonomous vehicles,'' in \emph{2018 IEEE Intelligent Vehicles Symposium (IV)}.\hskip 1em plus 0.5em minus 0.4em\relax IEEE, 2018, pp. 1--7.

\bibitem{xu2023diffscene}
C.~Xu, D.~Zhao, A.~Sangiovanni-Vincentelli, and B.~Li, ``Diffscene: Diffusion-based safety-critical scenario generation for autonomous vehicles,'' in \emph{The Second Workshop on New Frontiers in Adversarial Machine Learning}, 2023.

\bibitem{yin2021diverse}
Z.-H. Yin, L.~Sun, L.~Sun, M.~Tomizuka, and W.~Zhan, ``Diverse critical interaction generation for planning and planner evaluation,'' in \emph{2021 IEEE/RSJ International Conference on Intelligent Robots and Systems (IROS)}.\hskip 1em plus 0.5em minus 0.4em\relax IEEE, 2021, pp. 7036--7043.

\bibitem{shapley1953stochastic}
L.~S. Shapley, ``Stochastic games,'' \emph{Proceedings of the national academy of sciences (PNAS)}, vol.~39, no.~10, pp. 1095--1100, 1953.

\bibitem{mtr}
S.~Shi, L.~Jiang, D.~Dai, and B.~Schiele, ``Motion transformer with global intention localization and local movement refinement,'' \emph{Advances in Neural Information Processing Systems (Neurips)}, vol.~35, pp. 6531--6543, 2022.

\bibitem{qcnet}
Z.~Zhou, J.~Wang, Y.-H. Li, and Y.-K. Huang, ``Query-centric trajectory prediction,'' in \emph{Proceedings of the IEEE/CVF Conference on Computer Vision and Pattern Recognition (CVPR)}, 2023.

\bibitem{nuscenes}
H.~Caesar, V.~Bankiti, A.~H. Lang, S.~Vora, V.~E. Liong, Q.~Xu, A.~Krishnan, Y.~Pan, G.~Baldan, and O.~Beijbom, ``nuscenes: A multimodal dataset for autonomous driving,'' in \emph{Proceedings of the IEEE/CVF conference on computer vision and pattern recognition (CVPR)}, 2020, pp. 11\,621--11\,631.

\bibitem{womd}
P.~Sun, H.~Kretzschmar, X.~Dotiwalla, A.~Chouard, V.~Patnaik, P.~Tsui, J.~Guo, Y.~Zhou, Y.~Chai, B.~Caine, \emph{et~al.}, ``Scalability in perception for autonomous driving: Waymo open dataset,'' in \emph{Proceedings of the IEEE/CVF conference on computer vision and pattern recognition (CVPR)}, 2020, pp. 2446--2454.

\bibitem{metadrive}
Q.~Li, Z.~Peng, L.~Feng, Q.~Zhang, Z.~Xue, and B.~Zhou, ``Metadrive: Composing diverse driving scenarios for generalizable reinforcement learning,'' \emph{IEEE transactions on pattern analysis and machine intelligence}, vol.~45, no.~3, pp. 3461--3475, 2022.

\end{thebibliography}

\end{document}